\documentclass[conference]{IEEEtran}
\usepackage{amsthm}
\usepackage{amssymb}
\usepackage{mathtools}
\usepackage{bm}
\usepackage{nicefrac}
\makeatletter
\newcommand*{\T}{{\mathpalette\@transpose{}} }
\newcommand*{\@transpose}[2]{\raisebox{\depth}{$\m@th#1\intercal$}}
\makeatother
\DeclarePairedDelimiterX{\norm}[1]{\lVert}{\rVert_{2}}{#1}
\DeclareMathOperator*{\argmax}{arg\,max}

\usepackage{booktabs}

\usepackage[table,usenames,dvipsnames]{xcolor}
\definecolor{red}{HTML}{F07178}
\definecolor{green}{HTML}{62DE84}
\definecolor{blue}{HTML}{75A1FF}
\definecolor{orange}{HTML}{E55A1C}
\definecolor{yellow}{HTML}{FFCB6B}
\definecolor{gray}{HTML}{676E95}
\definecolor{white}{HTML}{FFFEFE}

\usepackage{tikz}
\usetikzlibrary{mindmap}
\usepackage{pgfgantt}
\usepackage[width=1cm,filledcolor=green,emptycolor=red]{progressbar}

\usepackage[numbers,sort&compress,square,comma]{natbib}  % \citet and \citep
\usepackage[colorlinks=true,urlcolor=red,citecolor=red]{hyperref}
\usepackage{cleveref}  % \cref must be loaded after hyperref and amsmath
\usepackage{epigraph}  % quotation
\usepackage{balance}

\usepackage{pifont}
\usepackage{listings}
\lstdefinestyle{mystyle}{
    commentstyle=\color{NavyBlue},
    keywordstyle=\color{Mulberry},
    numberstyle=\tiny\color{gray},
    stringstyle=\color{OliveGreen},
    basicstyle=\ttfamily\footnotesize,
    breaklines=true,
    captionpos=b,
    keepspaces=true,
    numbers=left,
    numbersep=5pt,
    frame=single,
    framexleftmargin=1.5em,
    xleftmargin=2em,
    showspaces=false,
    showstringspaces=false,
    showtabs=false,
    tabsize=2
}
\usepackage{graphicx}
\usepackage[caption=false,font=footnotesize]{subfig}

\usepackage[noend]{algpseudocode}
\usepackage{algorithm}
\usepackage{setspace}

\begin{document}
\title{Multi-Objective and Model-Predictive Tree Search for Spatiotemporal Informative Planning}
\author{
  \authorblockN{Weizhe Chen and Lantao Liu}
  \authorblockA{Indiana University, Bloomington, IN, USA, 47408\\
  Emails: \{chenweiz, lantao\}@iu.edu}
}

\maketitle
\begin{abstract}
    Adaptive sampling and planning in robotic environmental monitoring are challenging when the target environmental process varies over space and time.
    The underlying environmental dynamics require the planning module to integrate future environmental changes so that action decisions made earlier do not quickly become outdated.
    We propose a Monte Carlo tree search method which not only well balances the environment exploration and exploitation in space, but also catches up to the temporal environmental dynamics.
    This is achieved by incorporating multi-objective optimization and a look-ahead model-predictive rewarding mechanism.
    We show that by allowing the robot to leverage the simulated and predicted spatiotemporal environmental process, the proposed informative planning approach achieves a superior performance after comparing with other baseline methods in terms of the root mean square error of the environment model and the distance to the ground truth.
\end{abstract}
\IEEEpeerreviewmaketitle
%%%%%%%%%%%%%%%%%%%%%%%%%%%%%%%%%%%%%%%%%%%%%%%%%%%%%%%%%%%%%%%%%%%%%%%%%%%%%%
\section{Introduction and Related Work}
Environmental monitoring allows humans to obtain a good understanding of the underlying environmental state which oftentimes is spatially and temporally dynamic.
Capturing and utilizing the environmental dynamics are vital.
For example, when an oil spill happens in oceans,  the search-and-rescue crew need to estimate the oil diffusion and movement by integrating environmental information such as ocean currents, surface winds, and temperature/weather to best approximate the spill spreading behavior.
Autonomous robotic systems are increasingly being utilized as scientific data gathering tools, because they are efficient, economical, and able to monitor the environment that is too dangerous, or impossible, for humans to tackle~\cite{dunbabin2012robots}.
In robotic environmental monitoring, the mobile robot needs to build a ``distribution map'' for the environmental attribute of interest, based on which it is able to plan the ``best'' sampling trajectory.
Most existing literature uses Gaussian processes~(GPs) to construct the distribution map and uses adaptive sampling and planning methods to navigate the robot~\cite{lim2016adaptive,singh2009nonmyopic}.
GPs not only provide an estimate of the environmental attribute but also quantify the estimation uncertainty.
Given the estimation uncertainty, many information metrics, including but not limited to entropy, variance reduction, and mutual information can be derived.
Such information-theoretic planning framework is also called {\em informative planning}~\cite{meliou2007nonmyopic, hollinger2014sampling, ma2018data, chen2019pareto}, and it can effectively guide the mobile robot to the areas with the largest amount of information (or uncertainty reduction), leading to a good spatial coverage even within a short time window.

However, in the aforementioned environmental monitoring scenarios, good spatial coverage is insufficient to achieve satisfying modeling performance. The robot needs to explore the environment to discover the areas with high concentration (namely, \textit{hotspots}) first, and then visit different areas with differing efforts to obtain the most valuable samples while catching up with the spatiotemporal environmental dynamics.

\begin{figure}[tbp]
    \centering
    \includegraphics[width=0.8\linewidth]{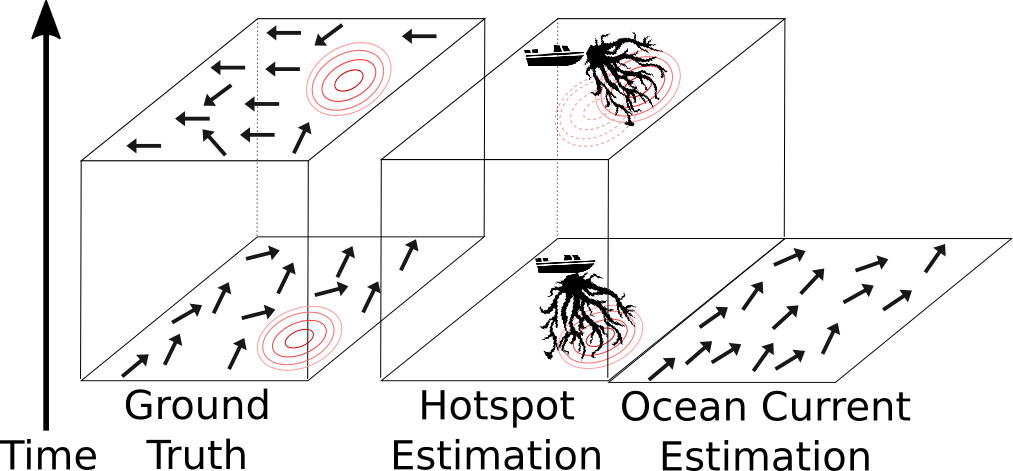}
    \caption{Illustration of current estimation based Monte Carlo tree search (MCTS) and model-predictive tree search in an oil spill monitoring task.
        A large volume of  spilled oil usually ends into multiple area patches on the surface which keep moving with eddies, see \href{https://oceanservice.noaa.gov/facts/oil-spill-trajectory.html}{the oil spill trajectory from NOAA} for example.
        The first column of this picture shows the unknown ground truth, where the arrows represent a vector field of the ocean currents, and the contour lines indicate a hotspot (patch) of spilled oil.
        In the second column, the bottom left image shows the robot's current estimation of the environment and the MCTS.
        The model-predictive tree search is carried out in the robot's prediction (upper layer) which comes from its current estimation of the oil density and the (imperfect) ocean current forecast.
    Model-predictive tree search encourages the robot to visit the area where the oil \textit{is likely to exist} while current estimation based MCTS guides the robot to where the oil \textit{has ever been}.}\label{fig:pull_fig}
\end{figure}

Recently, Monte Carlo tree search (MCTS) has gained great popularity as an online planning method.
It has been applied to spatiotemporal monitoring~\cite{marchant2014sequential}, planetary exploration mission~\cite{arora2017multi}, environmental exploration~\cite{corah2017efficient}, and active object recognition~\cite{best2018dec,patten2018monte}.
MCTS reveals a significant advantage for improving the performance of game-playing agents, especially in the game of Go~\cite{silver2016mastering}, based on perfect simulating environments with well-defined rules.
However, existing MCTS planners in robotics applications are typically based on the robot's current estimation of the environment without considering much on the (possibly highly) dynamic nature (see Fig.\,\ref{fig:pull_fig}).
This motivates us to develop an informative planning method that incorporates environmental prediction model where environmental dynamics are integrated with the MCTS working process.

Predicting the environmental process over space and time is a challenging task.
Nevertheless, there is a type of spatiotemporal environmental process whose dynamics are mainly affected by the motion of the environmental medium (e.g., fluid, air) in which it is located.
Fortunately, for these types of dynamic processes, simulation modeling is able to roughly estimate the possible trajectory of the feature of interest.
To name a few, such phenomena include spilled oil driven by ocean currents, wildfire affected by wind, and air pollution propagated by the prevailing wind.
Thanks to the efforts in these domains, we normally have access to some simulators that provide an (imperfect) estimation of the environmental process given current wind field or ocean currents data.
For example, General NOAA Operational Modeling Environment~(GNOME)~\cite{beegle2001general} is the modeling tool that predicts how wind, currents, and other processes might move and spread the spilled oil.
Fire Area Simulator (FARSITE) \cite{finney1998farsite} computes wildfire growth and behavior using wind, fuel moistures, and topography data.
An operational algal bloom forecast system is able to forecast the expansion of harmful algal blooms using a combination of satellite imaging and wind prediction~\cite{stumpf2009skill}.

This paper presents a multi-objective variant of MCTS with a simulator that enables the robot to incorporate its prediction of environmental state based on prior or related information.
The proposed method also allows the robot to handle potentially competing objectives such as exploration versus exploitation.
In addition, the algorithm can return a planning solution even if it is interrupted prematurely, making it suitable for online planning and \textit{in-sit} decision-making.

\section{PRELIMINARY}
\subsection{Gaussian Process Regression in Spatial Modeling}\label{sec: gp}
In the spatial modeling task, given a training set $\mathcal{D}$ of $n$ observations, $\mathcal{D} = \{(\bm{x}_i, y_i) | i = 1, \dots, n\}$, where $\bm{x}$ denotes an input location of dimension $2$ (or $3$) and $y$ denotes a scalar output or target, we want to infer the relationship between input and output.
Samples from the real world are assumed to be noisy, i.e. $y = f(\bm{x}) + \epsilon,$ where $f(\cdot)$ is the unknown function of interest and $\epsilon$ is the i.i.d. Gaussian noise with zero mean and variance $\sigma_n^2$, $\epsilon \sim \mathcal{N}(0, \sigma_n^2)$.

We place a Gaussian process (GP) prior over the unknown function, $f(\bm{x}) \sim \mathcal{GP}\left(\mu(\bm{x}), k(\bm{x}, \bm{x}'\right)$, where $\mu(\bm{x})$ is the mean function and $k(\bm{x}, \bm{x}')$ is the covariance function or kernel function.
The mean function is usually specified as zero and the data will be standardized when using GP.
Therefore, the estimation of GP is determined by the kernel function and the training data.
Let $X$ be a design matrix containing $n$ training inputs, $\bm{y}$ be the aggregated targets, and $X_*$ be the testing points corresponding to the function values $\bm{f}_*$.
The predictive distribution of GP regression is $\bm{f}_*|X, \bm{y}, X_* \sim \mathcal{N}(\bar{\bm{f}}_*\text{cov}(\bm{f}_*))$, where
\begin{align}
    \bar{\bm{f}}_* &= K_{X_*, X}K_X^{-1}
    \bm{y},\label{eq: post_mean}\\
    \text{cov}(\bm{f}_*) &= K_{X_*, X_*}
    - K_{X_*, X}K_X^{-1} K_{X, X_*}.
    \label{eq: post_cov}
\end{align}
Here we used the shorthand $K_X = [K_{X, X} + \sigma_n^2 I]$.
$K_{X_*, X}, K_{X_*, X_*}, \text{and} K_{X, X}$ are the covariance matrices.
The element $(i, j)$ in each matrix is calculated in a component-wise fashion: ${K_{X, X'}}{(i, j)} = k(x_i, x'_j)$.

The covariance function usually contains some adjustable parameters, which are called \textit{hyper-parameters}.
A widely adopted covariance function is the \textit{squared exponential kernel},
\begin{equation*}
    k_{SE}(\bm{x}, \bm{x'}) = \sigma_f^2 \exp{\left(-\frac{1}{2l^2}||\bm{x} - \bm{x'}||^2\right)},
\end{equation*}
which has hyper-parameters $\sigma_f^2$ and $l$.
The optimal set of parameters $\bm{\theta}^* = \{\sigma_f^2, l, \sigma_n^2\}$ is found by maximizing the \textit{log marginal likelihood} $\log p(\bm{y}|X, \bm{\theta})$.

Based on the posterior covariance (Eq.~\eqref{eq: post_cov}), the information metrics, such as entropy, mutual information, and variance reduction, can be calculated.
Recently, most planning algorithms for robotic environmental monitoring belong to \textit{informative planning} which aims to find the trajectory maximizing one of these information gain metrics.

\section{Problem Formulation}
Maximizing information gain tends to encourage exploration, which is not enough for catching up with the environmental dynamics.
As the environment is constantly changing, if the robot does not exploit the discovered features (e.g. hotspots) in time, the information contained in the collected data will ``expire'' because the environment may have changed significantly.
Since exploration and exploitation are two incompatible objectives, the problem falls within the spectrum of multi-objective optimization.

In the general case, multi-objective informative planning requires solving the following maximization problem:
\begin{align}
    \bm{a}^* = \argmax_{\bm{a} \in \mathcal{A}} \left\{f_1(\bm{a}),
    f_2(\bm{a}), \dots, f_{D}(\bm{a}) \right\}\quad
    \text{ s.t. } C_{\bm{a}} \leq B \notag
\end{align}
where $\bm{a}$ is a sequence of actions, $\mathcal{A}$ is the space of possible action sequences, $B$ is a budget (e.g. time, energy, memory, or number of iterations), $C_{\bm{a}}$ is the cost of actions, and $f_d(\bm{a}), d = 1, \dots, D$ are the objective functions defining which types of behaviors should the robot exhibit.
At least one of the $f_d(\bm{a})$ should be the information gain objective.

\section{METHODOLOGY}
\begin{figure}[tbp]
    \centering
    \includegraphics[width=0.8\linewidth]{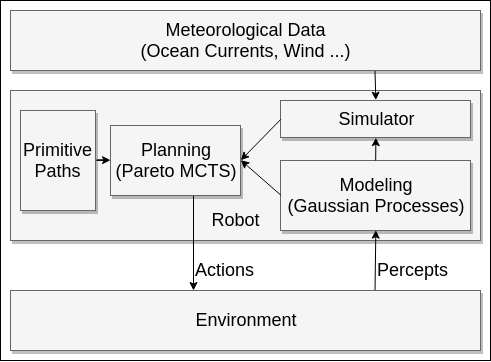}
    \caption{Block diagram of the proposed system.}
    \label{fig:system_diagram}
\end{figure}

Our objective is to monitor a time-varying phenomenon using an autonomous vehicle.
The robot needs to  build a ``distribution map'' based on the data collected for the environmental attribute of interest and plan the best sampling
path for the next step.
A big challenge in spatiotemporal monitoring is that, if the sampling path is planned based on the robot's estimation of the current
environmental state, it becomes obsolete and thus sub-optimal when the environment changes.
To address this, we proposed to unify the online informative planning (tree searching) mechanism with a forecast of the environmental state following estimated environmental dynamics.

The system, shown in Fig.\,\ref{fig:system_diagram}, consists of three main steps:
\begin{itemize} [leftmargin=13pt]
    \item \textit{Modeling.} As the robot travels along a path, it continuously collects data of the environment attribute that we are interested in.
        These newly acquired sample data are used to refine its estimation of the environment.
        Modeling is not the focus of this work, thus we employ the commonly adopted Gaussian process regression with the squared exponential kernel as described in Section~\ref{sec: gp}.
    \item \textit{Prediction.} Combining the estimation of the environment with the available ocean current or wind data, the robot predicts the possible moving path of the target by performing a forecast in its simulator.
        Details are presented in Section~\ref{sec: imagine}.
    \item \textit{Planning.} Given the Gaussian process and the prediction component, the robot searches a path to follow using Pareto MCTS.
        This is discussed in Section~\ref{sec: pareto_mcts}.
\end{itemize}

\subsection{Environment Prediction}\label{sec: imagine}
The \textit{prediction} essentially refers to the simulation of possible environmental states in the future based on meteorological data, such as ocean currents and wind field, as well as the current estimation of the environmental state.
Incorporating the prediction component is achievable because the robot can request the nowcast and forecast of the environmental conditions~(e.g., oceanographic information from Operational Forecast System and wind data from Global Forecast System).
Also, some off-the-shelf simulators are available for modeling the motion of the target in the ocean currents or wind fields, such as the trajectory of the oil spill~\cite{beegle2001general}, propagation of wildfire~\cite{finney1998farsite}, and expansion of harmful algal blooms~\cite{stumpf2009skill}.

Formally, let $X$ be the matrix containing all possible sampling locations and $\bm{v}_i^t(X), i = 1, \dots, V$ be $V$ vector fields representing different kinds of meteorological data at time $t$.
If we denote the current time as $t_0$ and the current estimation from GP (Eq.~\eqref{eq: post_mean}) as $\bm{f}_\text{predict}^{t_0}(X)$, then the environment simulator is a function that takes $\bm{f}_\text{predict}^{t_0}(X)$ and $\bm{v}_i^{t_0}(X), i = 1, \dots, V$ as inputs, and returns the prediction of the environment for the next time step $\bm{f}_\text{predict}^{t_{1}}(X)$.
Repeated use of the simulator allows us to obtain a series of predictions of the environment $\bm{f}_\text{predict}^{t_0}(X), \bm{f}_\text{predict}^{t_1}(X), \dots$.
These environment predictions will be part of the objective function for the planning module, guiding the robot to where hotspot may appear.
We will discuss the objective functions in section~\ref{sec: obj_func}.

\subsection{Primitive Paths}
\begin{figure}[t]
    \centering
    \includegraphics[width=0.4\linewidth]{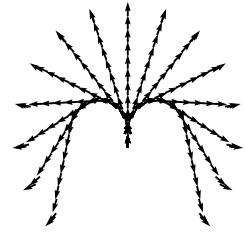}
    \caption{Illustration of robot's primitive paths. The arrows indicate the sampling positions and orientations of the robot as it moves along the path.}
    \label{fig:primitive_path}
\end{figure}

In this work, the robot is equipped with a pre-computed path generator which provides a set of primitive paths for every possible position.
Each path consists of a series of sampling points~(see~Fig.\,\ref{fig:primitive_path}).
Note that we will use the term \textit{action} and \textit{primitive path} interchangeably, but they mean the same thing.
The primitive paths are generated as follows:
\begin{itemize} [leftmargin=13pt]
    \item With some pre-defined radius around the robot, we obtain a given number of points (called endpoints) that evenly divide the circle. (Points that directly behind the robot might be discarded.)
    \item Connect the robot's current position with the endpoints using Dubin's model~\cite{dubins1957curves}.
        We assume the vehicle moves at constant speed in the plane and can execute turns with a minimum turning radius.
        Given the position and heading of the vehicle, Dubin's model gives the feasible shortest path joining these two oriented points.
    \item Environmental sampling data is then collected along a Dubin's path once it is chosen.
\end{itemize}

\subsection{Objective Functions}\label{sec: obj_func}
Although our proposed approach is applicable to the multi-objective problem, we will focus on the bi-objective case for an illustrative purpose.
Take the hotspot monitoring task as an example, two behaviors of the robot are expected in this task.
First, the robot needs to collect as much information about the environment as possible to determine the high-value areas~(i.e., hotspots).
Second, it also needs to visit these important areas more frequently in order to get better modeling performance.
To encourage these two behaviors, we need to define two objective functions~(or reward functions).

Given the number of look-ahead steps $\tau$, the hotspot-seeking reward function at $t_0$ is defined as the average of look-ahead prediction of the sampling points along a path at time $t_{\tau}$.
Suppose each primitive path $a$ has $m$ sampling points $X_a$, then the reward of taking action $a$ at time $t_0$ is given by
\begin{equation}
    f_{\text{hotspot}}^{t_0}(a) = \frac{1}{m}\sum
    \bm{f}_\text{predict}^{t_{\tau}}(X_a).  %,
    \label{eq: hotspot_reward}
\end{equation}
This objective function is a bridge between environment prediction and planning algorithms, encouraging the robot to visit where the target value will be higher in the future.

For the information gain reward function, we use the \textit{average reduction in variance}~(ARV)~\cite{binney2013optimizing}, because our method does not require the
objective function to have special properties, such as \textit{submodularity}~\cite{singh2009nonmyopic}, and it is handy to calculate.
Given the prior covariance matrix $K_{\text{prior}}$ and the posterior covariance matrix $K_{\text{post}}$ for $m$ sampling points of the GP, the ARV objective function is defined as
\begin{equation}
    f_{ARV}(a) = \frac{1}{m}[\text{Trace}(K_{\text{prior}}) -
    \text{Trace}(K_{\text{post}})].
    \label{eq: info_reward}
\end{equation}

Reward function Eq.~\eqref{eq: hotspot_reward} encourages the robot to the high-value areas and Eq.~\eqref{eq: info_reward} guides the robot to its unknown or uncertain regions.
The planning algorithm aims to choose a sequence of actions $\bm{a}$ that maximize these two reward functions,
\begin{align}
    \bm{a}^* = \argmax_{\bm{a} \in \mathcal{A}} \left\{
        f_{\text{hotspot}}(\bm{a}),
        f_{ARV}(\bm{a})
    \right\}
    \text{ s.t. } C_{\bm{a}} \leq B
    \label{eq: planning_problem}
\end{align}

Since the GP and the environment prediction are updated as the robot collects data, the reward functions are constantly changing.
Hence, we need an online planner that can adapt to the changing reward.
In addition, the planner should also consider the impact of the action on future results when choosing the current best action~(namely, \textit{nonmyopic} planning~\cite{singh2009nonmyopic}).

\subsection{Pareto Monte Carlo Tree Search}\label{sec: pareto_mcts}
\begin{figure}[tbp]
    \centering
    \subfloat[Symmetric Tree]{
        \resizebox{0.45\linewidth}{!}{
            \includegraphics[width=\linewidth,trim={7cm 7cm 7cm 7cm},clip]{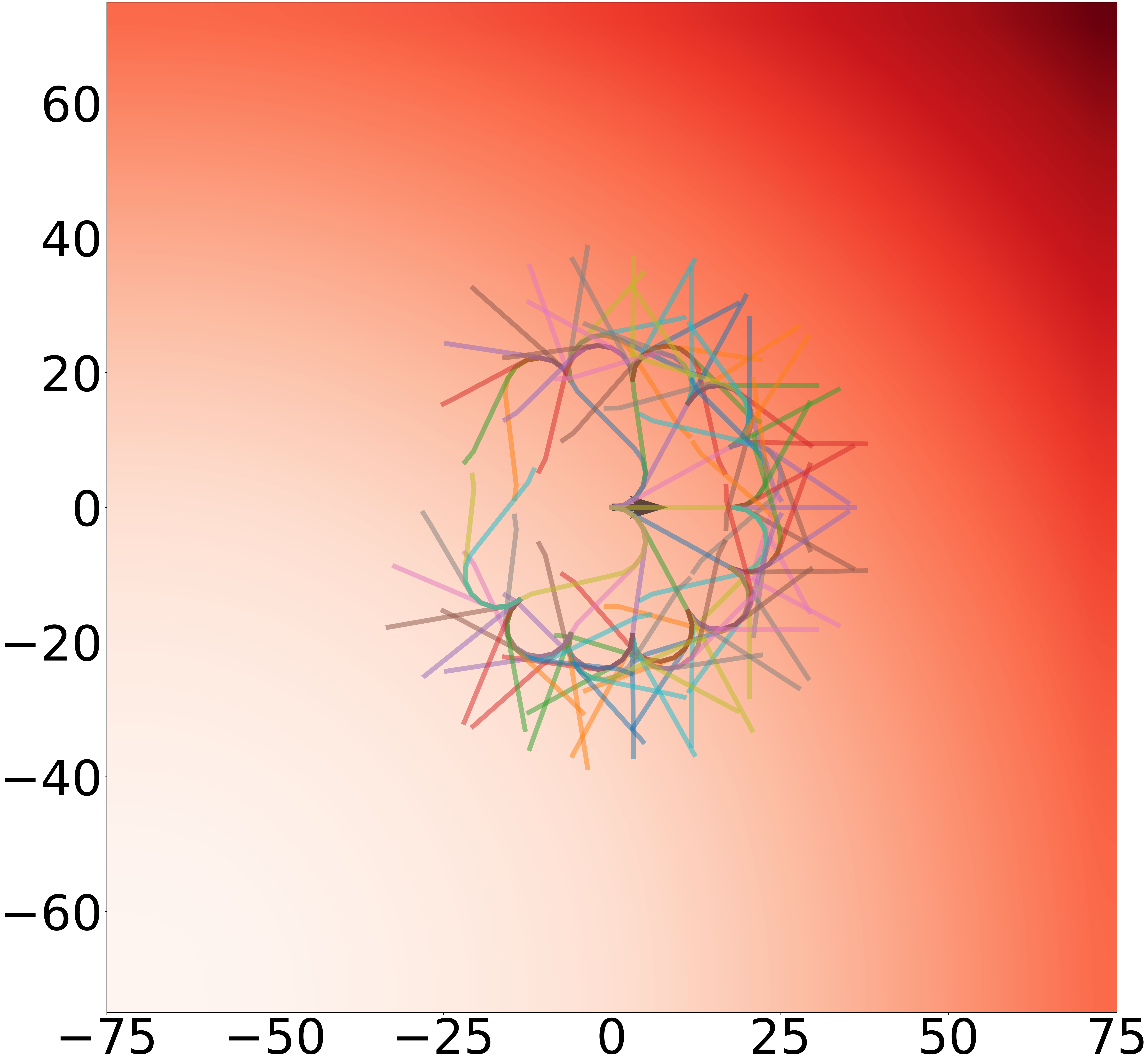}
        }
    }
    \subfloat[Asymmetric Tree]{
        \resizebox{0.45\linewidth}{!}{
            \includegraphics[width=\linewidth,trim={7cm 7cm 7cm 7cm},clip]{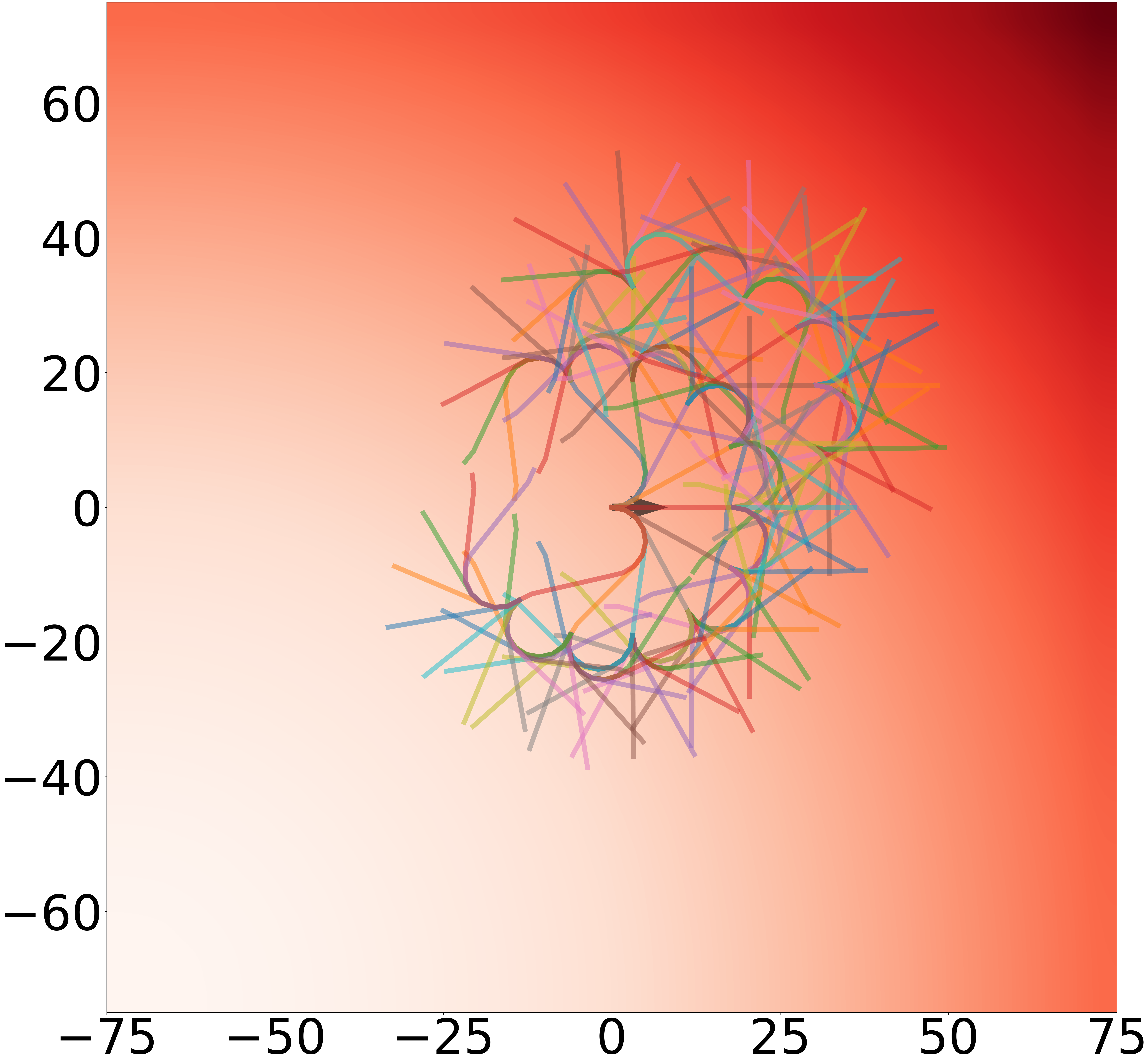}
        }
    }
    \caption{Illustration of a symmetric search tree and an asymmetric search tree with the same number of tree nodes. The robot is represented as the arrow. Red and white correspond to low and high reward, respectively. Pareto Monte Carlo tree search estimates the expected reward of each child node through Monte Carlo sampling, and puts more search efforts on the parts that are more likely to yield the best solution.}\label{fig:asymmetric}
\end{figure}

\begin{figure*}[t]
    \centering
    \includegraphics[width=0.7\linewidth]{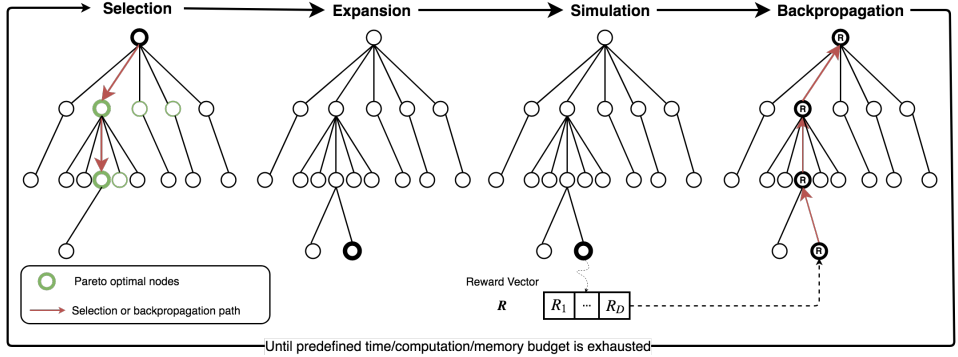}
    \caption{Illustration of Pareto Monte Carlo Tree Search main steps.}
    \label{fig:pareto_mcts}
\end{figure*}

One straightforward solution to Eq.~\eqref{eq: planning_problem} is to build a fixed-depth search tree that evaluates all  possible combinations of actions, then execute the first best action, and then replan, i.e., {\em receding-horizon tree search}.
However, the computational complexity of this method is prohibitive.
The Pareto Monte Carlo tree search (Pareto MCTS) alleviates this burden by building an asymmetric search tree which allocates its search time to the most important parts of the tree, as shown in Fig.\,\ref{fig:asymmetric}.

The tree is built incrementally through four steps: \textit{selection}, \textit{expansion}, \textit{simulation}, and \textit{backpropagation}~\cite{browne2012survey}.
The main steps of Pareto MCTS are depicted in Fig.\,\ref{fig:pareto_mcts}.
During each iteration, the algorithm descends through the tree (red arrows in the selection step) and a new node is attached to the tree (bold node in the expansion step), where each node represents a sampling location and each edge denotes an action leading to a subsequent location.
Starting from the location represented by the newly added node, we take some random actions and calculate the cumulative reward vector of taking these actions using the reward functions.
The cumulative reward vector will be utilized to update the estimation of expected reward stored in each node along the backpropagation path~(red arrows).

The selection step chooses an expandable node in the tree.
An expandable node has at least one action that has not yet been tried during the search.
To find an expandable node in a best-first search manner, we start from the root node and recursively apply a child node selection policy until an expandable node is reached.
We proposed an extension of the upper confidence bound~(UCB) policy~\cite{auer2002finite,kocsis2006bandit} for selecting the best next child in the multi-objective case~(described below).
In the expansion step, we randomly choose an untried action from the expandable node, calculate the robot's location after executing this primitive path, and build
a new child node according to this action and location.

During the simulation phase, the reward of the recently attached node is estimated by taking random actions and evaluating the cumulative reward of these actions until a pre-specified number of iterations is reached.
The estimated reward will be saved as a statistic in the node in order to bias future tree searches.
For our problem, the reward functions for evaluating the actions are defined as Eq.~\eqref{eq: hotspot_reward} and \eqref{eq: info_reward}.
This also implies that the reward here is not a scalar but a vector.
The standard MCTS assumes that the reward is a scalar, so special care should be taken during the selection, simulation, and backpropagation steps (reward is not used in the expansion step).
When arithmetic operations on the reward are needed, we perform element-wise operations.
For example, in the simulation and backpropagation steps, we need to calculate the cumulative reward.
This is done by adding up the corresponding elements in the reward vector.
Since the modification of the node selection policy is non-trivial, we will discuss it separately later.

The last step is the backpropagation where the operation of updating the statistics of all nodes along the path from the newly attached node to the root of the tree is performed.
Specifically, we increase the number of visits of all nodes along the path by one and add the reward evaluated in the simulation step to the cumulative reward of each node.

We repeat these four steps until the maximum number of iterations is exceeded or a given time limit is exhausted.
Finally, the algorithm returns the best action corresponding to the optimal child of the root node in terms of expected reward.
Pareto MCTS is an anytime algorithm which is able to return a valid action at any moment during the searching process.
A better action can be found if it is allowed to run for a longer time.

For the critical node selection step in Pareto MCTS, one needs to balance exploitation of the currently most promising child nodes and exploration of alternatives which may turn out to be a superior choice at a later time.
We develop a selection policy that is able to handle the vector-form reward and balances exploration and exploitation following the \textit{optimism in the face of uncertainty} principle~\cite{auer2002finite}.

\textbf{Pareto optimality}: In order to choose the best child node, we need to define a metric first.
In the case of scalar reward, the most promising node is simply the one with the highest expected reward.
However, given two reward vectors, it is not obvious to tell which one is superior to the other.
This is where the Pareto optimal set comes into play.

In a multi-objective optimization problem, there does not typically exist a solution that maximizes all objective functions simultaneously.
Therefore, the goal is to find the \textit{Pareto optimal set} instead, or the so-called \textit{Pareto front}.
The solutions in the Pareto front cannot be improved for any objective without hurting other objectives.
Formally, let $\bar{r}_d^k$ be the expected reward corresponding to the $d$-th objective function of child node $k$.
A node $i$ is said to (Pareto) \textit{dominate} another node $j$ if $\bar{r}_d^i \geq \bar{r}_d^j$ for all $d = 1, \dots, D$, and $\bar{r}_d^i > \bar{r}_d^j$ for at least one $d\in \{1, 2, \dots, D\}$.
A node is called Pareto optimal node if there does not exist another node that dominates it.
The set of Pareto optimal nodes is called Pareto front or Pareto optimal set.
It is worth mentioning that any node in this set is considered equally best.

\textbf{Node selection policy}: In the node selection step, we evaluate a score vector for each child and then choose the best child in terms of Pareto optimality.

First, we calculate the following upper confidence bound vector for each child node~\cite{chen2019pareto}:
\begin{equation}\label{eq: PUCB}
    \bm{U}(k) = \frac{\bm{r}^k}{n_k} + \sqrt{\frac{4\ln{n} + \ln{D}}{2 n_k}},
\end{equation}
where $\bm{r}^k$ is the cumulative reward vector stored in node $k$, $n_k$ is the number of times child $k$ has been visited, $D$ is the number of dimensions of the reward, and $n = \sum_{k=1}^K n_k$ is the number of times the parent node has been visited.
Eq.\,\eqref{eq: PUCB} is a combination of the empirical expected reward and a confidence interval.
The first term prefers the recently discovered most promising child whereas the second term emphasizes exploration of alternatives.
When the number of visits of the $k$-th node remains the same while that of other nodes increases, the second term will become larger, making node $k$ more likely to be selected.

Second, we build a Pareto optimal set based on the resulting $\bm{U}(k), k = 1, \dots, K$.
Alg.\,\ref{alg:pareto} outlines the main steps for computing the Pareto optimal set for a bi-objective case.
See~\cite{kung1975finding} for more general multi-objective cases.

{\algrenewcommand\textproc{}
    \begin{algorithm}[t!]
        \setstretch{1.1}\small%
        \caption{\textbf{Computing Bi-Objective Pareto Front}.}\label{alg:pareto}%
        \begin{algorithmic}[1]
            \Procedure{$\mathtt{compute\_pareto\_front}$}{$\mathbf{U}$}
            \State \Comment{$\mathbf{U}$ is a $K\times{2}$ matrix consists of upper confidence vectors}
            \State $\mathbf{U}_\text{sort} \leftarrow$ sort $\mathbf{U}$ w.r.t. the first column in descending order
            \State $\mathbf{u}^{\star}\leftarrow\mathbf{U}_\text{sorted}[0,:]$\Comment{get the first row vector}
            \State $\mathbb{P}\leftarrow\{\text{node corresponds to }\mathbf{u}^{\star}\}$\Comment{add to Pareto optimal set}
            \For{$i=1,\dots,K-1$}
            \If{$\mathbf{u}^{\star}[1]\leq\mathbf{U}_\text{sort}[i,1]$}\Comment{Compare the second dim.}
            \State $\mathbf{u}^{\star}\leftarrow\mathbf{U}_\text{sort}[i,1]$
            \State $\mathbb{P}\leftarrow\mathbb{P}\cup\{\text{node corresponds to }\mathbf{u}^{\star}\}$
            \EndIf
            \EndFor
            \State \textbf{return} $\mathbb{P}$
            \EndProcedure
        \end{algorithmic}
    \end{algorithm}
}

Finally, the best child is chosen from the Pareto optimal set uniformly at random.
The reason for choosing the best child randomly is that the Pareto optimal nodes are considered equally optimal if no preference information is specified.

\section{EVALUATION RESULTS}
In order to verify the advantages of incorporating multiple objectives and environment prediction into planning, we compared the UCT~\cite{kocsis2006bandit}, Pareto MCTS, and the proposed look-ahead based multi-objective tree search in the task of monitoring a spatially and temporally dynamic hotspot.

\subsection{Experiment Setup}
The task of the robot is to monitor a hotspot that passively moved by its fluid medium.
One can think about this as monitoring the oil spill or harmful algal bloom in the ocean.
The robot is equipped with $11$ primitive paths, each of which is $2.5$ m long.
It collects data in a two-dimensional $150\text{ m } \times 150\text{ m }$ environment with range
($x_{\min}$: $-75$, $x_{\max}$: $75$, $y_{\min}$: $-75$, $y_{\max}$: $75$).
The hotspot is specified by a Gaussian distribution with initial mean $\bm{m} = [0, -50]^\top$.
The robot starts from $[-20, -40]$ with no prior knowledge of the environment, so it needs to explore the environment first to find the hotspot.

The input of GP only includes the spatial dimension and the spatiotemporal correlation is not modeled in this work, thus we assume that the environment is static in a short time period.
The GP will be reset and the robot is allowed to take $5$ actions at each time step.
For the $s$-th ($s = 0, \dots, 4$) action $a_s$ at time $t_0$, the hotspot-seeking reward function is defined as the average s-step look-ahead prediction
\begin{equation}
    f_{\text{hotspot}}^{t_0}(a_s) = \frac{1}{m}\sum
    f_\text{predict}^{t_s}(X_*).
    \label{eq: hotspot_reward_experiment}
\end{equation}

We also assume that a forecast of ocean currents with near future is available although noisy~\cite{liu2018solution}.
At each location $(x, y)$ and time $t$, the predicted ocean current vector is a noisy version of the true ocean current:
\begin{equation}
    \bm{v}^t_{\text{predict}} = \bm{v}^t_{\text{true}} + \bm{\epsilon}
    =
    \begin{pmatrix}
        5\cos{\left(2\pi\frac{t\text{ mod }60}{60}\right)} + \epsilon_x\\
        5\sin{\left(2\pi\frac{t\text{ mod }60}{60}\right)} + \epsilon_y
    \end{pmatrix},
    \label{eq: predicted_current}
\end{equation}
where $\epsilon_x, \epsilon_y$ are independent Gaussian noises drawn from $\mathcal{N}(0, 0.2)$.

The Gaussian distribution is linearly moved (disturbed) by the ocean current vector:
\begin{equation}
    \bm{m}_{t+1} = \bm{m}_t + \bm{v}^t_{\text{true}}
    \label{eq: relationship}
\end{equation}

In the environment prediction component, we first find out the locations where the GP estimations are greater than the median of all estimations.
Then, these locations are moved by the predicted ocean current vector $\bm{v}^t_{\text{predict}}$ as in Eq.~\eqref{eq: relationship}.

In the three compared planning algorithms, UCT only uses the ARV objective function and selects the child node maximizing the following upper confidence bound in the selection step:
\begin{equation}
    U(k) = \frac{r^k}{n_k} + \sqrt{\frac{2\ln{n}}{n_k}},
    \label{eq: UCB}
\end{equation}
where $r^k$ is the scalar cumulative reward stored in node $k$ and other variables have the same meaning as the variables in Eq.~\eqref{eq: PUCB}.
Note that when $D = 1$, Eq.~\eqref{eq: PUCB} degenerates into Eq.~\eqref{eq: UCB}.
Our proposed Eq.~\eqref{eq: PUCB} is a multidimensional extension of Eq.~\eqref{eq: UCB}.
Pareto MCTS differs from the proposed method in that it only uses the current estimation of the environment in its hotspot-seeking reward function, and does not involve any environmental model prediction.
All three versions of MCTS use the same parameters: $200$ maximum number of tree search iterations and $2$ random actions in the simulation step.
The hyper-parameters of GP are updated by maximizing the log marginal likelihood in an online manner.

\subsection{Results}
\begin{figure*}[tb]
    \centering
    \includegraphics[width=1\linewidth, height=6cm]{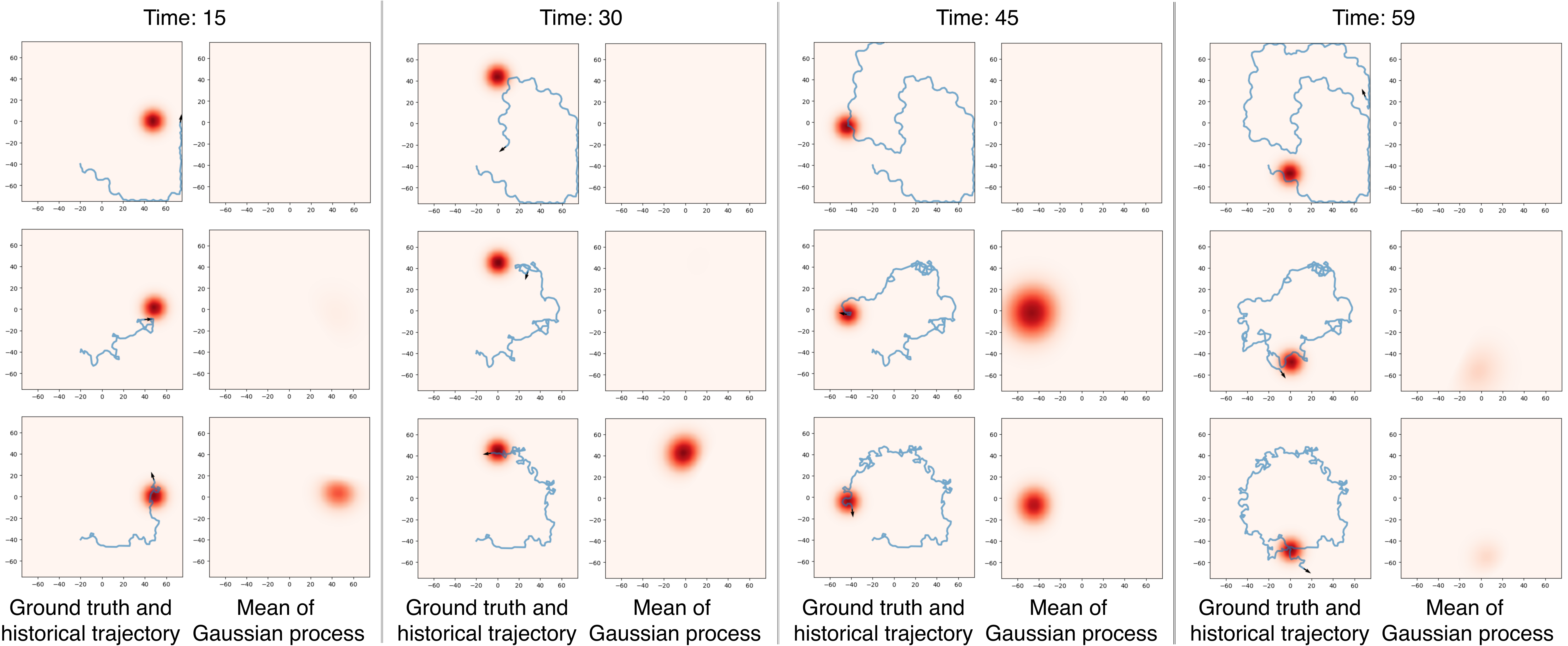}
    \caption{Snapshots of the robot's behaviors in the hotspot monitoring task using UCT (first row), Pareto MCTS (second row) and model-predictive tree search (third row) at time $15, 30, 45$, and $59$. The arrows indicate the positions and orientations of the robot. Blue lines are the historical trajectories. For each time step, the left image is the ground truth of the environment, and the right one is robot's current estimation of the environment.}\label{fig:snapshots}
\end{figure*}

Fig.\,\ref{fig:snapshots} presents the snapshots of the robot's behaviors when using UCT, Pareto MCTS, and model-predictive tree search.
At the beginning of the mission, all the three algorithms guide the robot to explore the environment, looking for the hotspot.
This can be seen from the historical trajectories of the robot, because the trajectory will appear to be stretched when the robot is exploring, and it will be circling during exploitation.
When the hotspot is found, the three algorithms show different characteristics.

UCT with only information gain objective function ignores the hotspot and continues exploring~(See the first row of Fig.\,\ref{fig:snapshots}), because less information will be gained from a sampling point once it has been observed.
Without exploitation at a finer level, we cannot identify the hotspot feature in the mean of GP.
This explains why it is insufficient to use only the information gain objective in spatiotemporal monitoring.

Pareto MCTS tends to visit the high-value area more frequently once it has been discovered (second row of Fig.\,\ref{fig:snapshots}).
However, the robot will easily lose the target and needs to search for it again~(second row of time $30$ and $45$).
As a comparison, the proposed method not only better tracks the hotspot, but also guides the robot to the area that the hotspot may exist in the next time (third row of time $30$).

The left pictures of the fourth column show the complete historical paths.
The path planned by the proposed method is the most consistent with the ground-truth underlying hotspot moving path~(a circle).
It can be seen that the robot moderately wandered around the hotspot to collect more valuable samples.

\begin{figure}[tbp]
    \centering
    \includegraphics[width=1\linewidth]{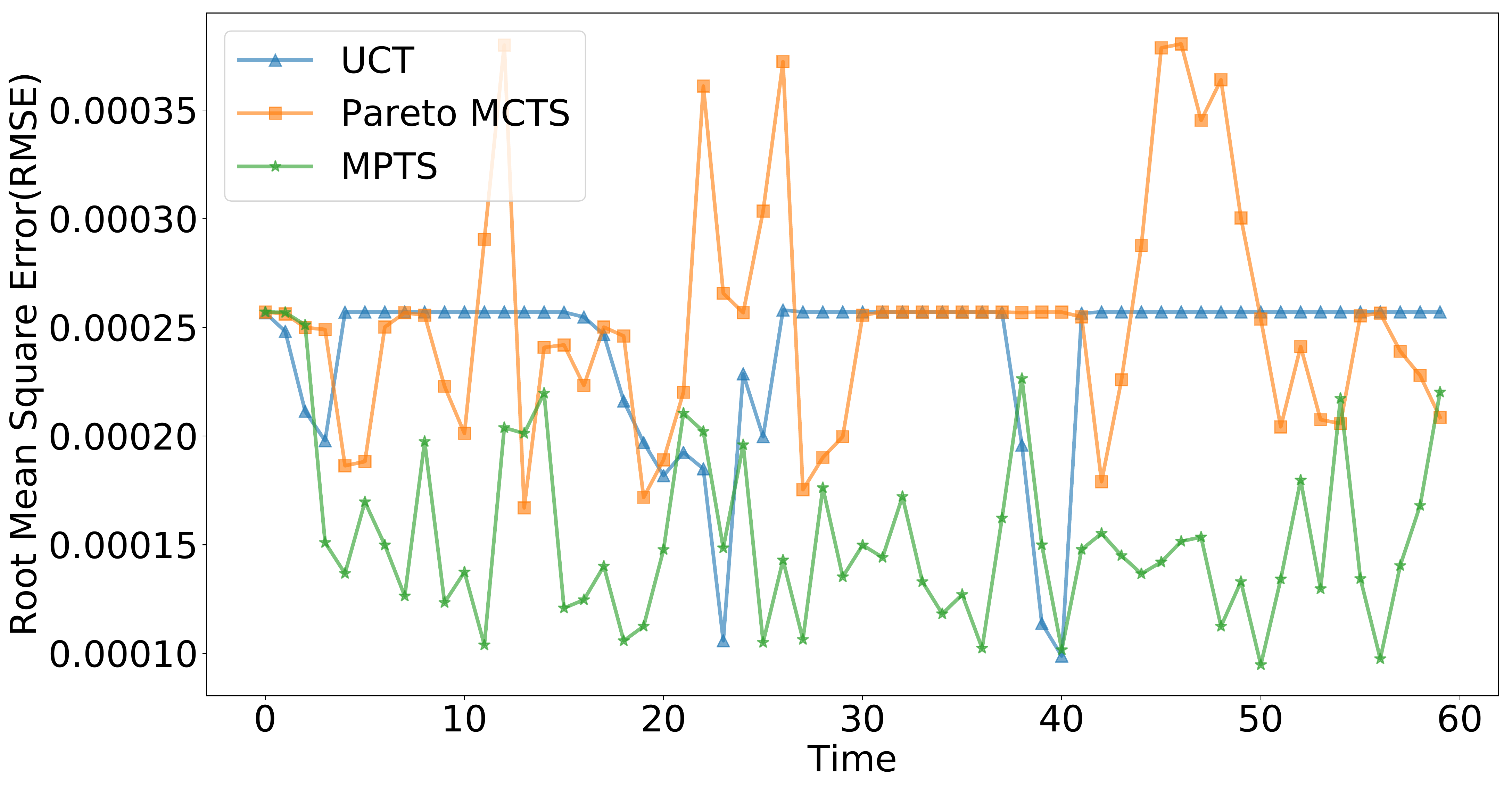}
    \caption{The root mean square error (RMSE) of the three version of MCTS.}\label{fig:rmse}
\end{figure}

\begin{figure}[tbp]
    \centering
    \includegraphics[width=.95\linewidth]{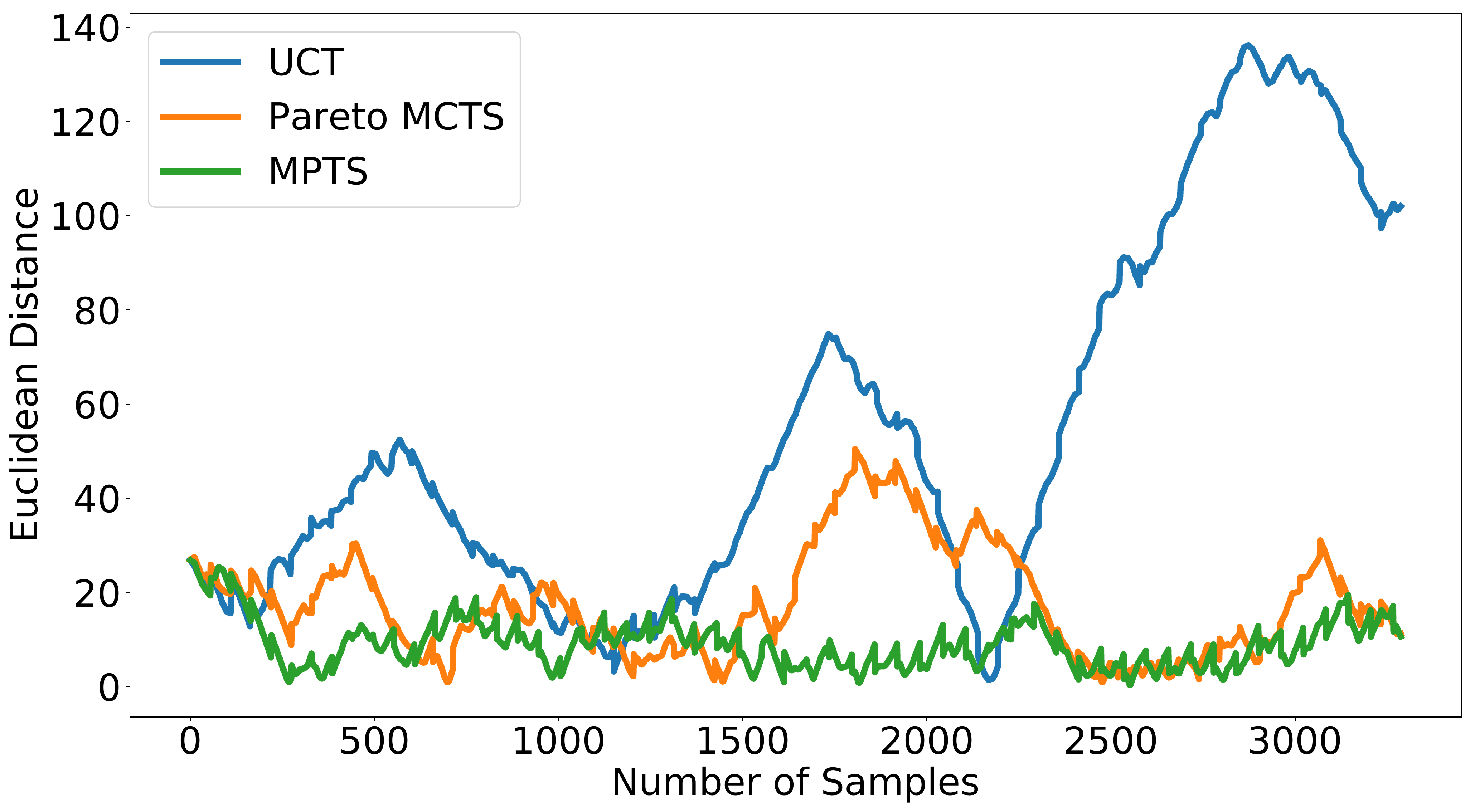}
    \caption{Distance from the robot's position to the center of the hotspot.}
    \label{fig:hotspot_distance}
\end{figure}

These behavioral characteristics directly affect the final modeling performance.
From the right image of each column in Fig.\,\ref{fig:snapshots}, we can see that UCT fails to model the hotspot, Pareto MCTS overestimates the range of the hotspot, and the model-predictive tree search provides a good estimate for the underlying hotspot.
We also quantitatively compare the root mean square error~(RMSE) in Fig.\,\ref{fig:rmse} and distance to the center of the hotspot in Fig.\,\ref{fig:hotspot_distance}.
When the robot does not collect any sample in the hotspot, the GP predicts the value of all locations as $0$, resulting in $0.000255$ RMSE.
The robot using UCT did not find the hotspot most of the time, but occasionally encountered hotspot during the exploration process~(Fig.\,\ref{fig:hotspot_distance}).
As can be seen from Fig.\,\ref{fig:rmse} and Fig.\,\ref{fig:hotspot_distance}, Pareto MCTS lost the hotspot during the time period $[30, 40]$, but most of time it was able to collect data around hotspot.
Since it is often unable to catch up with the dynamics of the environment, the collected samples do not represent the underlying hotspot well, leading to unsatisfactory RMSE.
The RMSE of the model-predictive tree search drops drastically and fluctuates at low values, which reveals that the proposed method can quickly find the hotspot, then track it and exploit it.

\section{CONCLUSION and FUTURE WORK}
This paper presents an informative planning framework for robotic monitoring of spatiotemporal environments.
According to the nowcast and forecast of the environmental model, the robot is able to infer the future changes of the target and plan its sampling actions accordingly.
Within this framework, we propose a sampling-based multi-objective tree search planning algorithm that allows the robot to handle multiple potentially conflicting objectives, such as exploration and exploitation.
The results of a hotspot evolution monitoring experiment show that our proposed method enables the robot to adapt to the time-varying environment and balance exploration and exploitation behaviors, achieving smaller modeling error and closer distance to the hotspot center.

%%%%%%%%%%%%%%%%%%%%%%%%%%%%%%%%%%%%%%%%%%%%%%%%%%%%%%%%%%%%%%%%%%%%%%%%%%%%%%%
\balance
\bibliographystyle{abbrvnat}
\bibliography{references}
\end{document}